\documentclass[11pt]{article}
\usepackage{coling2016}
\usepackage{times}
\usepackage{url}
\usepackage{latexsym}
\usepackage{multirow}
\usepackage{amsmath,amsfonts,amssymb}
\usepackage{graphicx,subfigure}
\usepackage{hyperref}
\usepackage{paralist}

\title{Modelling Sentence Pairs with Tree-structured Attentive Encoder}

\author{ Yao Zhou, Cong Liu$^\ast$ \and Yan Pan\\ 
School of Data and Computer Science, Sun Yat-sen University \\
  {\tt yoosan.zhou@gmail.com}\\ \tt \{liucong3, panyan5\}@mail.sysu.edu.cn}

\date{}
\begin{document}
\maketitle
\begin{abstract}
We describe an attentive encoder that combines tree-structured recursive neural networks and sequential recurrent neural networks for modelling sentence pairs. Since existing attentive models exert attention on the sequential structure, we propose a way to incorporate attention into the tree topology. Specially, given a pair of sentences, our attentive encoder uses the representation of one sentence, which generated via an RNN, to guide the structural encoding of the other sentence on the dependency parse tree. We evaluate the proposed attentive encoder on three tasks: semantic similarity, paraphrase identification and true-false question selection. Experimental results show that our encoder outperforms all baselines and achieves state-of-the-art results on two tasks.
\end{abstract}

\section{Introduction}
\label{intro}
\blfootnote{
\hspace{-0.65cm}
$^\ast$ indicates the corresponding author. \\
Code is available at \url{https://github.com/yoosan/sentpair} \\
This work is licensed under a Creative Commons Attribution 4.0 International License. License details: \url{http://creativecommons.org/licenses/by/4.0/} \\
}

Modelling a sentence pair is to score two pieces of sentences in terms of their semantic relationship. The applications include measuring the semantic relatedness of two sentences~\cite{marelli2014semeval}, recognizing the textual entailment~\cite{bowman2015large} between the premise and hypothesis sentences, paraphrase identification~\cite{he2015multi}, answer selection and query ranking~\cite{yin2015abcnn} etc.

The approach of modelling a sentence pair based on neural networks usually consist of two steps. First, a sentence encoder transforms each sentence into a vector representation. Second, a classifier receives two sentence representations as features to make the classification. The sentence encoder can be regarded as a semantic compositional function which maps a sequence of word vectors to a sentence vector. This compositional function takes a range of different forms, including (but not limited to) sequential recurrent neural networks (Seq-RNNs)~\cite{mikolov2012statistical}, tree-structured recursive neural networks (Tree-RNNs)~\cite{socher2014grounded,tai2015improved} and convolutional neural networks (CNNs)~\cite{kim2014convolutional}.

We introduce an approach that combines recursive neural networks and recurrent neural networks with the attention mechanism, which has been widely used in the sequence to sequence learning (\emph{seq2seq}) framework whose applications ranges from machine translation~\cite{bahdanau2014neural,luong2015effective}, text summarization~\cite{rush2015neural} to natural language conversation~\cite{shang2015neural} and other NLP tasks such as question answering~\cite{NIPS2015_5846,hermann2015teaching}, classification~\cite{rocktaschel2016reasoning,shimaoka2016attentive}. In the machine translation, the attention mechanism is used to learn the alignments between source words and target words in the decoding phase. More generally, we consider that the motivation of attention mechanism is to allow the model to attend over a set of elements with the intention of attaching different emphases to each element. We argue that the attention mechanism used in a tree-structured model is different from a sequential model. Our idea is inspired by Rockt{\"a}schel et al.~\shortcite{rocktaschel2016reasoning} and Hermann et al.~\shortcite{hermann2015teaching}. In this paper, we utilise the attention mechanism to select semantically more relevant child by the representation of one sentence learned by a Seq-RNNs, when constructing the head representation of the other sentence in the pair on a dependency tree. Since our model adopts the attention in the sentence encoding phase, we refer to it as an attentive encoder. In this work, we implement this attentive encoder with two architectures: tree-structured LSTM and tree-structured GRU.

We evaluate the proposed encoder on three sentence pair modelling tasks: semantic similarity on the SICK dataset, paraphrase identification on the MSRP dataset and true-false question selection on the AI2-8grade science questions dataset. Experimental results demonstrate that our attentive encoder is able to outperform all non-attentional counterparts and achieves the state-of-the-art performance on the SICK dataset and AI2-8grade dataset.

\section{Models}
\begin{figure}[t]
\centering
\includegraphics[scale=.62]{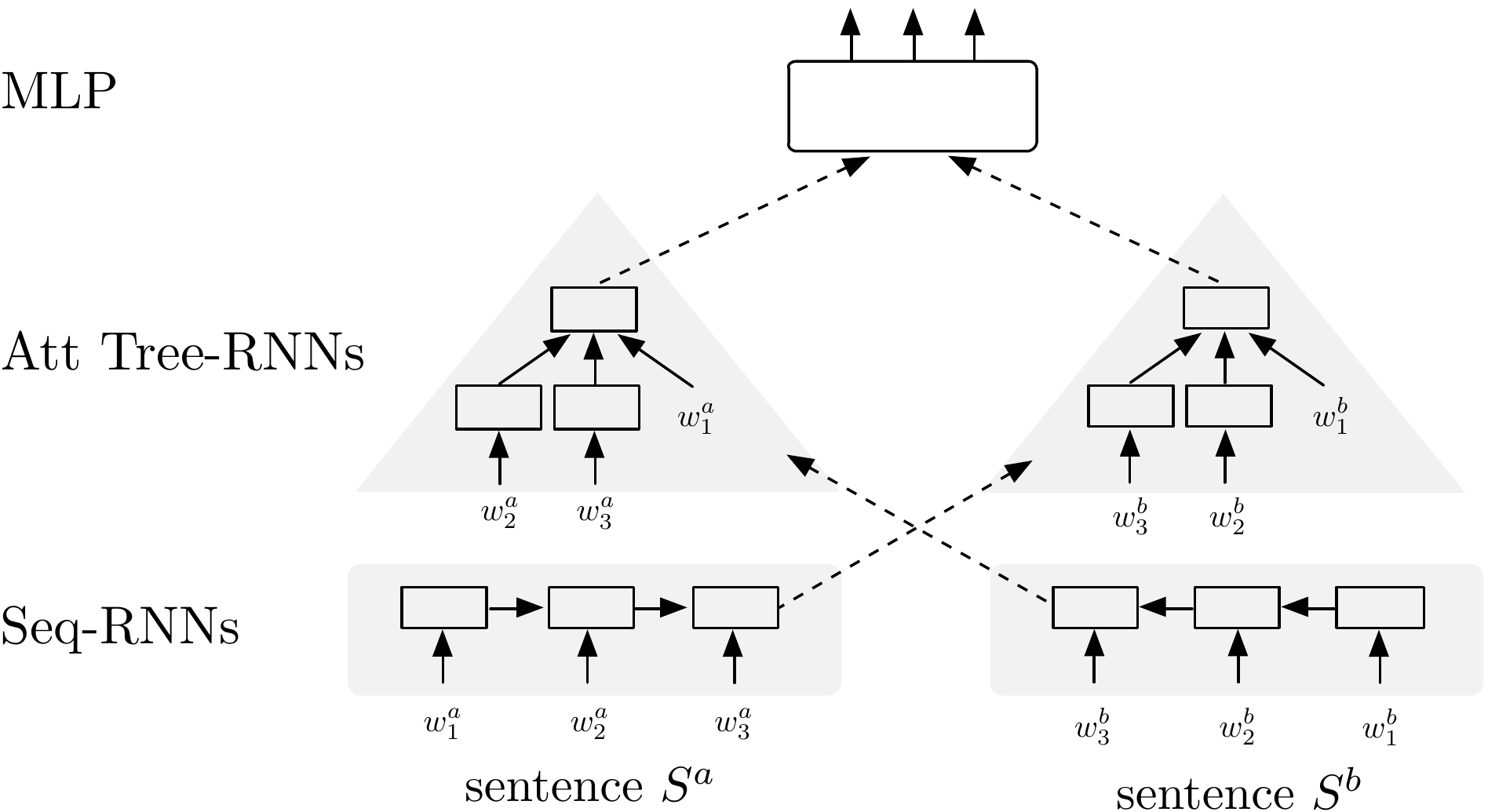} \caption{An overview of our tree-structured attentive encoder.}
\label{fig-model}
\end{figure}

Let's begin with a high-level discussion of our tree-structured attentive encoder. As shown in Figure~\ref{fig-model}, given a sentence pair ($S^{a}$, $S^{b}$), our goal is to score this sentence pair. Our tree-structured attentive model has two components. In the first component, a pair of sentences is fed to a Seq-RNNs, which encodes each sentence and results in a pair of sentence representations. In second component, the Attentive Tree-RNNs encodes a sentence again, aimed by the representation of the other sentence generated by the first component. Compared with the existing approaches of modelling sentence pairs, our attentive encoder consider not only the sentence itself but also the other sentence in the pair. Finally, the two sentence vectors produced by the second component are fed to the multilayer perceptron network to produce a distribution over possible values. These components will be detailed in the following sections.

\subsection{Seq-RNNs}
We first describe the RNN composer, which is the basic unit of Seq-RNNs. Given an input sequence of arbitrary length, an RNN composer iteratively computes a hidden state $h_t$ using the input vector $x_t$ and its previous hidden state $h_{t-1}$. In this paper, the input vector $x_t$ is a word vector of the $t$-th word in a sentence. The hidden state $h_t$ can be interpreted as a distributed representation of the sequence of tokens observed up to time $t$. Commonly, the RNN transition function is the following:
\begin{equation}
\label{equ-rnn}
h_t = \tanh(Wx_t + Uh_{t-1} + b)\\
\end{equation}

\begin{figure*}[t]
\centering
\subfigure[LSTM]{\label{figure-lstm} \includegraphics[scale=0.48]{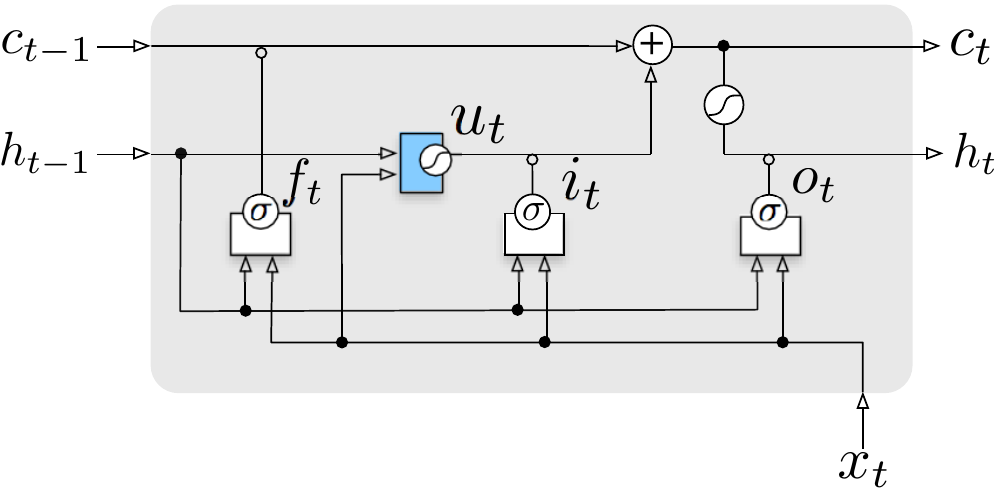}}
\subfigure[Child-Sum Tree-LSTM]{\label{figure-treelstm} \includegraphics[scale=0.51]{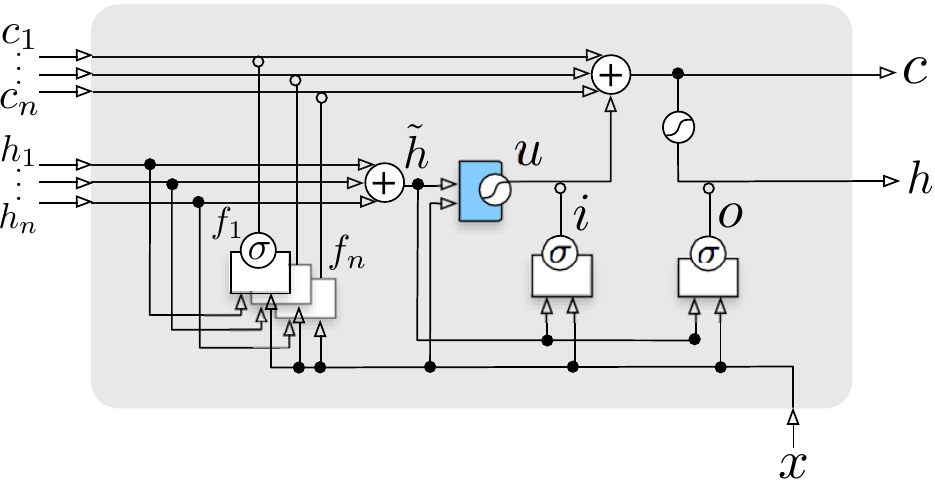}}
\subfigure[Attentive Child-Sum Tree-LSTM]{\label{figure-attlstm} \includegraphics[scale=0.52]{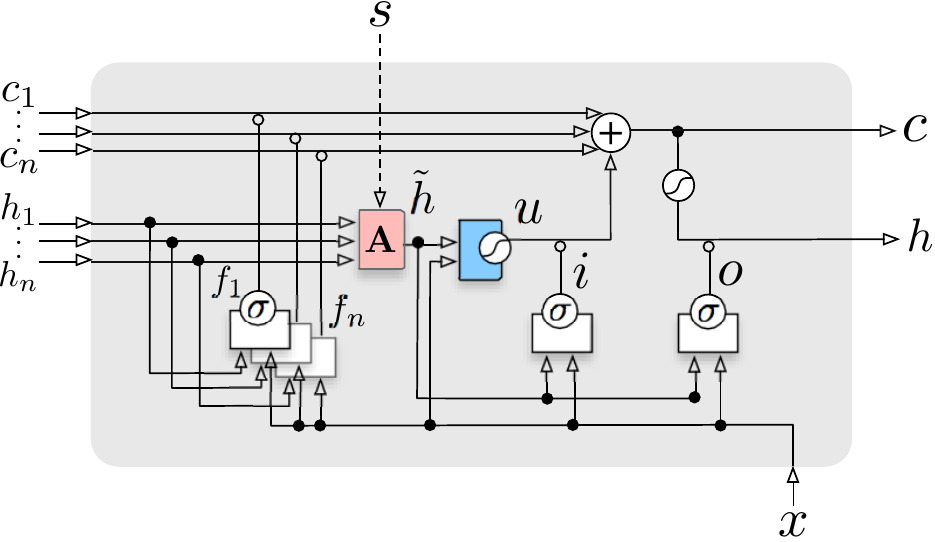}}
\subfigure[GRU]{\label{figure-gru} \includegraphics[scale=0.48]{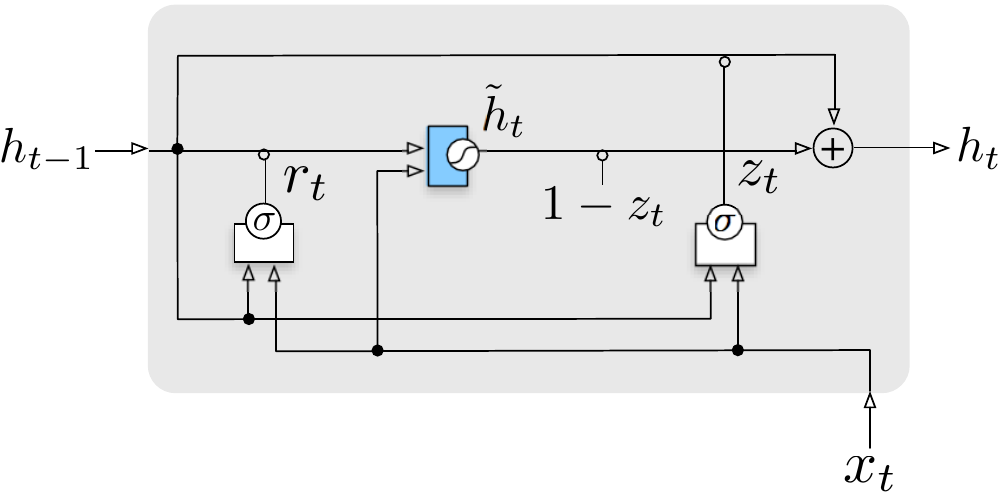}}
\subfigure[Child-Sum Tree-GRU]{\label{figure-treegru} \includegraphics[scale=0.51]{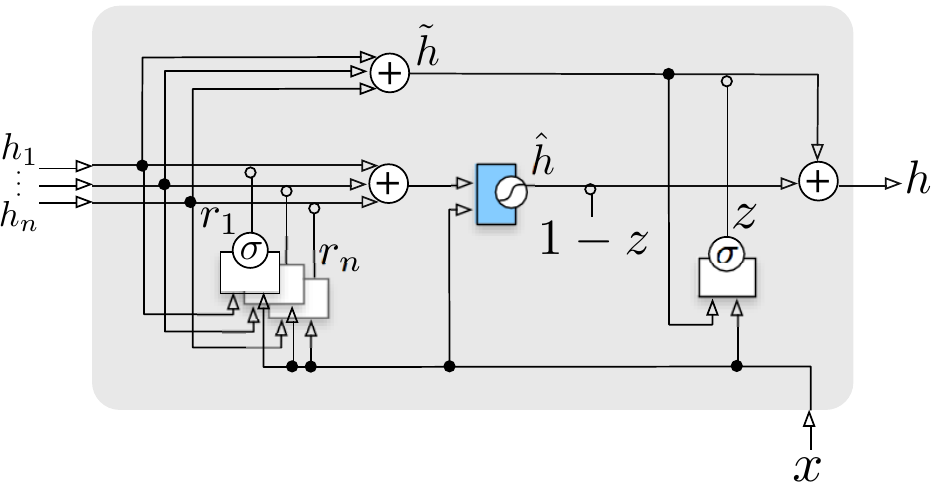}}
\subfigure[Attentive Child-Sum Tree-GRU]{\label{figure-attgru} \includegraphics[scale=0.51]{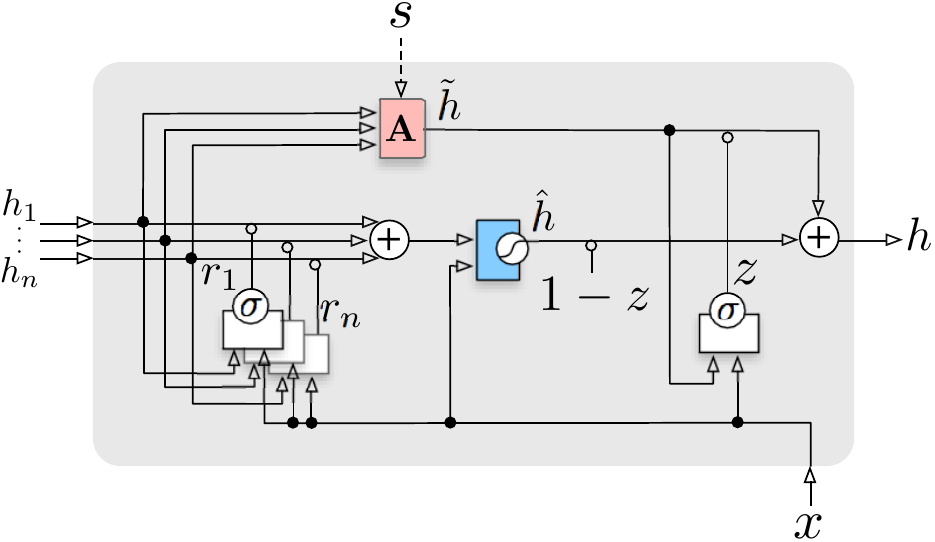}}
\caption{Illustrations of different recurrent architectures.}
\end{figure*}

We refer to the model that recursively apply the RNN composer to a sequence as the Seq-RNNs. Unfortunately, standard Seq-RNNs suffers from the problem that the gradients of the hidden states of earlier part of the sequence vanishes in long sequences~\cite{hochreiter1998vanishing}. Long Short-term Memory (LSTM)~\cite{hochreiter1997long} and Gated Recurrent Unit (GRU)~\cite{chung2014empirical} are two powerful and popular architectures that address this problem by introducing gates and memory. In this paper, we only show the illustrations of LSTM~(Figure \ref{figure-lstm}) and GRU~(Figure \ref{figure-gru}). The implementations of standard LSTM and GRU in this paper are same as~\cite{luong2015effective} and~\cite{chung2014empirical}. When we replace the standard RNN composer with LSTM or GRU, the Seq-RNNs becomes Seq-LSTMs or Seq-GRUs.

\subsection{Standard Tree-RNNs}
Compared with standard RNN composer, which computes its hidden state from the input at the current time step and the hidden state of previous time step, the Tree-RNN composer computes its hidden state from an input and the hidden states of arbitrarily many child units. We now describe the \emph{Child-Sum Tree-LSTM} and \emph{Child-Sum Tree-GRU} architectures which are formed by applying the \emph{Child-Sum} algorithm to LSTM and GRU respectively.

\paragraph{Child-Sum Tree-LSTM.} In this paper, the implementation of Child-Sum Tree-LSTM is same as~\cite{tai2015improved}. We consider that a Child-Sum Tree-LSTM composer contains two parts: the external part and internal part. The external part consists of the inputs and outputs, and the internal part is the controllers and memory of the composer. As shown in Figure~\ref{figure-treelstm}, the inputs of the composer are: a input vector $x$, multiple hidden states $h_1, h_2, \dots, h_n$ and  multiple memory cells $c_1, c_2, \dots, c_n$, where $n$ is the number of child units. The outputs consist of a memory cell $c$ and a hidden state $h$ which can be interpreted as the representation of a phrase. The internal part aims at controlling the flow of information by an input gate $i$, an output gate $o$ and multiple forget gates $f_1, f_2, \dots, f_n$. The gating mechanisms used in the Child-Sum Tree-LSTM are similar to sequential LSTM. Intuitively, the sum of children's hidden states $\tilde{h}$ is the previous hidden state, the forget gate $f_k$ controls the degree of memory kept from that of the child $k$, the input gate $i$ controls how much the internal input $u$ is updated and the output gate controls the exposure of internal memory $c$. We define the transition equations as follows:

\begin{align} \label{equation-treelstm}
\begin{array}{ll}
\tilde{h}= \sum_{1\leq k\leq n} h_{k}, ~~~~~& ~~~~~~~~i=\sigma(W^{(i)}x+U^{(i)}\tilde{h}+b^{(i)}),\\
o=\sigma(W^{(o)}x+U^{(o)}\tilde{h}+b^{(o)}), ~~~~~& ~~~~~~~~u=\tanh(W^{(u)}x+U^{(u)}\tilde{h}+b^{(u)}),\\
f_k=\sigma(W^{(f)}x+ U^{(f)}h_k + b^{(f)}), ~~~~~& ~~~~~~~~c=i \odot u + \sum_{1\leq k\leq n}f_k \odot c_k,\\
h= o\odot \tanh(c), \\
\end{array}
\end{align}

\paragraph{Child-Sum Tree-GRU.} The way of Child-Sum Tree-GRU extending to the standard GRU is similar to the way of Child-Sum Tree-LSTM extending to the standard LSTM. Since we only introduce the Child-Sum algorithm applied to the LSTM and GRU, in the following, we omit the ``Child-Sum" prefix of Child-Sum Tree-LSTM and Child-Sum Tree-GRU for simplicity. Compared with the Tree-LSTM, the Tree-GRU removes the memory cell $c$ and introduces an update gate $z$ and multiple reset gates $r_1, r_2, \dots, r_n$ that allow the composer to reset the hidden states of the child units. The candidate hidden state $\hat{h}$ is computed similarly to the standard RNN (Equation~\ref{equ-rnn}) and the update gate $z$ is used to control how much the previous hidden state $\tilde{h}$ and the candidate $\hat{h}$ should be passed. The transition equations of Tree-GRU are in the following:
\begin{align} 
\label{equation-treegru}
\begin{array}{ll}
\tilde{h}= \sum_{1\leq k\leq n} h_ {k}, & r_k = \sigma(W^{(r)}x + U^{(r)}h_{k} + b^{(r)}),\\
\hat{h} = \tanh(W^{(h)}x + U^{(h)} \sum_{k=1}^{n}r_{k} \odot h_{k}), & z = \sigma(W^{(z)}x + U^{(z)}\tilde{h} + b^{(z)}) \\
h = z \odot \tilde{h} + (1-z) \odot \hat{h},\\
\end{array}
\end{align}
where $\sigma$ denotes the sigmoid function and $\odot$ denotes element-wise multiplication.

We can easily apply the Child-Sum Tree-RNN to the dependency trees that have branching factors of arbitrary number and order-insensitive nodes. We refer to the model adopting the Tree-LSTM and Tree-GRU composer to the dependency tree as the \emph{Dependency Tree-LSTMs} and \emph{Dependency Tree-GRUs}. For simplicity, we also omit the prefix ``Dependency'' in the following sections.

\subsection{Attentive Tree-RNNs}
We now details how we extend the standard Tree-RNN. The idea that we incorporate the attention into the standard Tree-RNN comes from: (1) there will be semantic relevance between two sentences in the sentence pair modelling tasks; (2) the effect of semantic relevance could be implemented in the process of constructing the sentence representation by Tree-RNN where each child should be assigned a different weight; and (3) the attention mechanism is well suited for learning weights on a contextual collection where a guided vector is attending over.
\paragraph{Soft Attention Layer}
In this work, the attention mechanism is implemented by a \emph{soft attention layer} $A$. Given a collection of hidden states $h_1, h_2, \dots, h_{n}$ and an external vector $s$, the soft attention layer produce a weight $\alpha_k$ for each hidden state as well as a weighted vector $g$ via the Equations~\ref{equ-attention}:

\begin{align} 
\label{equ-attention}
\begin{array}{ll}
m_{k} = \tanh(W^{(m)}h_{k} + U^{(m)}s), ~~~~~& ~~~~~~~~\alpha_{k}=\frac{\exp(w^{\intercal}m_{k})}{\sum_{j=1}^{n}\exp(w^{\intercal}m_{j})}, \\
g=\sum_{1\leq k\leq n}\alpha_{k}h_{k},\\
\end{array}
\end{align}

\paragraph{Attentive Tree-LSTM and -GRU}

As illustrated in Figure~\ref{figure-attlstm} and Figure~\ref{figure-attgru}, when the attention layer is embedded to the standard Tree-LSTM and Tree-GRU composer, these composers become the Attentive Tree-LSTM and Attentive Tree-GRU. The attention layer (notated with $A$) receives the children's hidden states and an external vector $s$, producing a weighted representation $g$ (Equation~\ref{equ-attention}). In our implementation, the external vector $s$ is a vector representation of sentence learned by a Seq-RNNs. Specifically, if the tree composer is Attentive Tree-LSTM, then the Seq-RNNs is Seq-LSTMs. Instead of taking the sum of children's hidden states as the previous hidden state $\tilde{h}$ in the standard Tree-RNN, we compute a new hidden state by a transformation $\tilde{h} = \tanh(W^{(a)}g + b^{(a)})$. Similar to the standard Tree-RNN, the attentive composers can also be easily applied to dependency trees. We refer to the model which applies the Attentive Tree-RNN composer to the dependency tree as the Attentive (Dependency) Tree-RNNs.

\subsection{MLP}
\label{sec-mlp}
The multilayer perceptron network (MLP) receives a pair of vectors produced by the sentence encoder to compute a multinomial distribution over possible values. Given two sentence representations $h_{L}$ and $h_{R}$, we compute their component\-wise product $h_{L} \odot h_{R}$ and their absolute difference $|h_{L} - h_{R}|$. These features are also used by~Tai et al.\shortcite{tai2015improved}. We then compress these features into a low dimensional vector $h_s$, which is used to compute the probability distribution $\hat{p}_{\theta}$. The equations are the following:
\begin{align}
\begin{split}
&h_{\times}=h_L\odot h_R, ~~~~~h_{+} = |h_L - h_R |, \\
&h_s = \sigma (W^{(\times)}h_{\times} + W^{(+)}h_{+} + b^{(h)}), \\
&\hat{p}_{\theta} = \text{softmax}(W^{(p)}h_s + b^{(p)}),\\
\end{split}
\end{align}

\section{Experiments and Results}
\begin{table}[htpb]
\begin{center}
\begin{tabular}{|l|l|l|l|}
\hline 
\textbf{Config} & \textbf{Value}  & \textbf{Config} & \textbf{Value}\\ 
\hline
Word vectors & Glove~\cite{pennington2014glove} & Dims of word vectors & 300 \\
\hline
OOV word vectors & uniform(-0.05, 0.05) & Dims of hidden state & 150 \\
\hline
Learning rate & 0.05 & Batch size & 25 \\
\hline
Regularization & L2 with $\lambda = 10^{-4}$ & Dropout  rate & 0.5\\
\hline
Optim method & Adagrad~\cite{duchi2011adaptive} & Num of epoch & 10 \\
\hline
\end{tabular}
\end{center}
\caption{\label{table-config}The training configs. We use the 300D Glove vectors as the initial word vectors. Out-of-vocabulary words are initialized with a uniform distribution. The model parameters are regularized with a per-minibatch L2 regularization strength of $10^{-4}$. The dropout is used at the classifier with a dropout rate 0.5. All models are trained using Adagrad with a learning rate of 0.05. We train our models for 10 epochs, and pick the model that has the best results on the devlopment set to evaluate on the test set.}
\end{table}

\paragraph{Our baselines}In order to make a meaningful comparison between the sequential models, tree-structured models and attentive models, we present four baselines. They are:
\begin{inparaenum}[(i\upshape)]
 \item \textbf{Seq-LSTMs}, learning two sentence representations by the sequential LSTMs;
 \item \textbf{Seq-GRUs}, like Seq-LSTMs but using GRU composer; 
 \item \textbf{Tree-LSTMs}, learning two sentence representations by the Dependency Tree-LSTMs; and
 \item \textbf{Tree-GRUs}, like Tree-LSTMs but using Child-Sum Tree-GRU composer.
\end{inparaenum}
The two sentence representations are fed to the MLP to produce a probability distribution.
\subsection{Task 1: Semantic Similarity}
First we conduct our semantic similarity experiment on the Sentences Involving Compositional Knowledge(SICK) dataset~\cite{marelli2014semeval}\footnote{Dependency trees are parsed by the Stanford Parser package, \url{http://nlp.stanford.edu/software/lex-parser.html}}\footnote{Glove vectors are available at \url{http://nlp.stanford.edu/projects/glove/}}. This task is to predict a similarity score of a pair of sentences, based on human generated scores. The SICK dataset consists of 9927 sentence pairs with the split of 4500 training pairs, 500 development pairs and 4927 testing pairs. Each sentence pair is annotated with a similarity score ranging from 1 to 5. A high score indicates that the sentence pair is highly related. All sentences are derived from existing image and video annotation dataset. The evaluation metrics are Pearson's $r$, Spearman's $\rho$ and mean squared error (MSE).

Recall that the output of MLP (Section~\ref{sec-mlp}) is a probability distribution $\hat{p}_{\theta}$. Our goal in this task is to predict a similarity score of two sentences. Let $r^{\intercal} = [1, \dots, 5]$ be an integer vector, the similarity score $\hat{y}$ is computed by $\hat{y} = r^{\intercal} \hat{p}_{\theta}$. We take the same setup as~\cite{tai2015improved} that computes a target distribution $p$ as a function of prediction score $y$ given by:
\begin{equation*}
p_{i} = \left\{
\begin{array}{ll}
y - \lfloor{y}\rfloor, & i = \lfloor{y}\rfloor + 1 \\
\lfloor{y}\rfloor - y + 1, & i = \lfloor{y}\rfloor \\
0 & \text{otherwise}
\end{array} \right.
\end{equation*} The loss function of semantic similarity is the KL-divergence that measures the continuous distance between the predicted distribution $\hat{p}_{\theta}$ and the distribution of the ground truth $p$:
\begin{equation}
J(\theta) = \frac{1}{N}\sum_{k=1}^{N}\text{KL}(p^{(k)} \Big|\Big| \hat{p}_{\theta}^{(k)}) + \frac{\lambda}{2}||\theta||_{2}^{2}
\end{equation}
\begin{table}[t]
\begin{center}
\begin{tabular}{|l|c|c|c|}
\hline \bf Method & \bf{$r$}  & \bf{$\rho$} & \bf MSE \\ 
\hline
ECNU~\cite{zhao2014ecnu} & 0.8414 & - & - \\ 
Dependency Tree-LSTMs~\cite{tai2015improved} & 0.8676 & 0.8083 & 0.2532 \\
combine-skip+COCO~\cite{kiros2015skip} & 0.8655 & 0.7995 & 0.2561 \\
ConvNet~\cite{he2015multi} & 0.8686 & 0.8047 & 0.2606 \\
\hline
Seq-GRUs & 0.8595 & 0.7974 & 0.2689 \\
Seq-LSTMs & 0.8528 & 0.7911 & 0.2831\\
(Dependency) Tree-GRUs & 0.8672 & 0.8116 & 0.2573  \\
(Dependency) Tree-LSTMs(ours) & 0.8664 & 0.8068 & 0.2610 \\
\hline
\multicolumn{4}{|l|}{\bf +Attention} \\
Attentive (Dependency) Tree-GRUs & 0.8701 & 0.8085 & 0.2524\\
Attentive (Dependency) Tree-LSTMs & \bf 0.8730 & \bf 0.8117 & \bf 0.2426  \\
\hline
\end{tabular}
\end{center}
\caption{\label{results-similarity} Test set results on the SICK dataset. The first group is previous results, and remaining is ours.}
\end{table}

The results are summarized in Table~\ref{results-similarity}. We first compare our results against the previous results. ECNU~\cite{zhao2014ecnu}, the best result of SemEval 2014 submissions, achieves a 0.8414 $r$ score by a heavily feature-engineered approach. Kiros et al.~\shortcite{kiros2015skip} presents an unsupervised approach to learn the universal sentence vectors without depending on a specific task. Their Combine\textendash skip+COCO model improve the Pearson's $r$ to 0.8655, but a weakness is that their sentence vectors are high-dimensional vectors (2400D). Training the skip-thoughts vectors needs a lot of time and space. He et al. \shortcite{he2015multi} show the effectiveness of convolutional nets with the similarity measurement layer for modelling sentence similarity. Their ConvNet outperforms ECNU with +0.027 Pearson's $r$. We can observe that dependency Tree-LSTM, combine-skip+COCO and ConvNet almost achieve the same performance and our Attentive Tree-LSTMs outperforms these three methods around +0.005 points. Comparison to ECNU, our Attentive Tree-LSTMs gains an improvement of +0.032 and achieves the state-of-the-art performance. We find a phenomenon also appeared in~\cite{tai2015improved} that tree-structured models can outperform sequential counterparts. Comparison to the non-attentional baselines (such as Tree-LSTMs), the attention mechanism (such as Attentive Tree-LSTMs) gives us a boost of around +0.007.  All results highlight that our attentive Tree-RNNs are well suited for the semantic similarity task.
\subsection{Task 2: Paraphrase Identification}
The next task we evaluate is paraphrase identification on the Microsoft Research Paraphrase Corpus (MSRP)~\cite{dolan2004unsupervised}. Given two sentences, this task is to predict whether or not they are paraphrases. The dataset is collected from news sources and contains 5801 pairs of sentences, with 4076 for training and the remaining 1725 for testing. We randomly select 10\% of training set and use them as our dev set. This task is a binary classification task, therefore we report the accuracy and F1 score.

Since that the $\hat{p}_{\theta}$ indicates the distribution over the possible labels, we take $argmax(\hat{p}_{\theta})$ as the predicted label in the testing phase. The loss function for the binary classification is the \emph{binary cross-entropy}:
\begin{equation}
J(\theta) = -\frac{1}{N}\sum_{k=1}^{N}(y^{(k)}\text{log}\hat{p}_{\theta}^{(k)} + (1-y^{(k)})\text{log}(1-\hat{p}_{\theta}^{(k)})) + \frac{\lambda}{2}||\theta||_{2}^{2}
\end{equation}

\begin{table}
\centering
\scriptsize
\subtable{
\begin{tabular}{lccc}
\hline

\hline

\bf Method & \bf Acc(\%)  & \bf F1(\%) \\ 
\hline
Baseline~\cite{mihalcea2006corpus} & 65.4 & 75.3 \\
RAE~\cite{socher2011dynamic} & 76.8 & 83.6 \\
combine-skip+feats~\cite{kiros2015skip} & 75.8 & 83.0 \\
ABCNN-3~\cite{yin2015abcnn} & 78.9 & 84.8 \\
TF-KLD~\cite{ji2013discriminative} & \bf 80.4 & \bf 85.9 \\
\hline
Seq-GRUs & 71.8 & 80.2 \\
Seq-LSTMs & 71.7 & 80.6 \\
Tree-GRUs & 73.6 & 81.8 \\
Tree-LSTMs & 73.5 & 82.1\\
\multicolumn{3}{l}{\bf +Attention} \\
Attentive Tree-GRUs & 74.8 & 82.3 \\
Attentive Tree-LSTMs & \bf 75.8 & \bf 83.7 \\
\hline

\hline
\end{tabular}
}
\qquad
\subtable{
\begin{tabular}{lcc}
\hline

\hline

\bf Method & \bf Dev Acc(\%) & \bf Test Acc(\%)\\ 
\hline
RNN~\cite{baudis2016joint} & 38.1 & 36.1 \\
CNN~\cite{baudis2016joint} & 44.2 & 38.4 \\
RNN-CNN~\cite{baudis2016joint} & 43.9 & 37.6 \\
attn1511~\cite{baudis2016joint} & 38.4 & 35.8 \\
Ubu.RNN~\cite{baudis2016joint} & 49.4 & 44.1 \\
\hline
Seq-GRUs & 72.1 & 62.4 \\
Seq-LSTMs &  71.8 & 63.3\\
Tree-GRUs & 75.2 & 70.6 \\
Tree-LSTMs & 74.6 & 69.1 \\

\multicolumn{3}{l}{\bf +Attention} \\
Attentive Tree-GRUs & \bf 76.4 & 72.1\\
Attentive Tree-LSTMs & 76.2 & \bf 72.5\\
\hline

\hline
\end{tabular}
}
\caption{\label{results-2}The test results of paraphrase identification on the Microsoft Paraphrase Corpus (\textbf{Left}) and true-false selection on the AI2-8grade dataset (\textbf{Right}).}
\end{table}

Table~\ref{results-2} (\textbf{left}) presents our results on the MSRP dataset. The previous approaches are:
\begin{inparaenum}[(1\upshape)]
 \item Baseline, cosine similarity with tf-idf weighting;
 \item RAE, recursive autoencoder with dynamic pooling;
 \item combine-skip+feats, skip-thought vectors with features;
 \item ABCNN-3, attention-based convolutional nets; and
 \item TF-KLD, matrix factorization with supervised reweighting.
\end{inparaenum} First, all our models are able to outperform the baseline. We only compare our models with the neural networks-based approaches, including RAE and ABCNN-3 for a fair comparison. We find that our models do not prove to be very competitive. After a careful analysis, we conclude that the reasons are (1) our models are pure neural networks-based, we don't add any features to identify paraphrases while the other methods have used additional features; (2) The MLP is not very suitable in this task. We attempt to replace the MLP with the cosine distance and euclidean distance in our future work. Although our models have not yet matched the SOTA performance, we obtain an improvement of +2.3 accuracy by Attentive Tree-LSTMs when we incorporate the attention into the standard Tree-LSTM.
\subsection{Task 3: True-False Question Selection}
We last consider a challenging task: selecting true or false given a scientific question and its evidence. In this task, we use the AI2-8grade dataset built by~\cite{baudis2016joint}. This dataset is derived from the \emph{AI2 Elementary School Science Questions} released by Allen Institute. Each sentence pair consists of a hypothesis sentence processed by substituting the $wh$-word in the question by answer and its evidence sentence extracted from a collection of CK12 textbooks. The number of sample pairs in the training, development, and test set are 12689, 2483 and 11359 respectively. This dataset contains 626 words not appearing in \emph{Glove vectors}, most of which are named entities and scientific jargons.

The loss function is the same as the paraphrase identification since this task is also a binary classification task. We reports the accuracy on development set and test set shown in Table~\ref{results-2} (\textbf{right}). Since this dataset is a fresh and uncompleted dataset, we only compare our models with Baudis et al.~\shortcite{baudis2016joint} who have evaluated several models on it. Comparison to~\cite{baudis2016joint}, all of our models gain a significant improvement. Specially, our best result achieved by the Attentive Tree-LSTMs is higher than the best of ~\cite{baudis2016joint} by +28 percents. It is observed that tree-structured models are more competitive than the sequential counterparts. As we expected, the attentive models can outperform all non-attentional counterparts.

\begin{figure*}[t]
\centering
\subfigure[SICK dataset]{\label{len-sick} \includegraphics[scale=0.35]{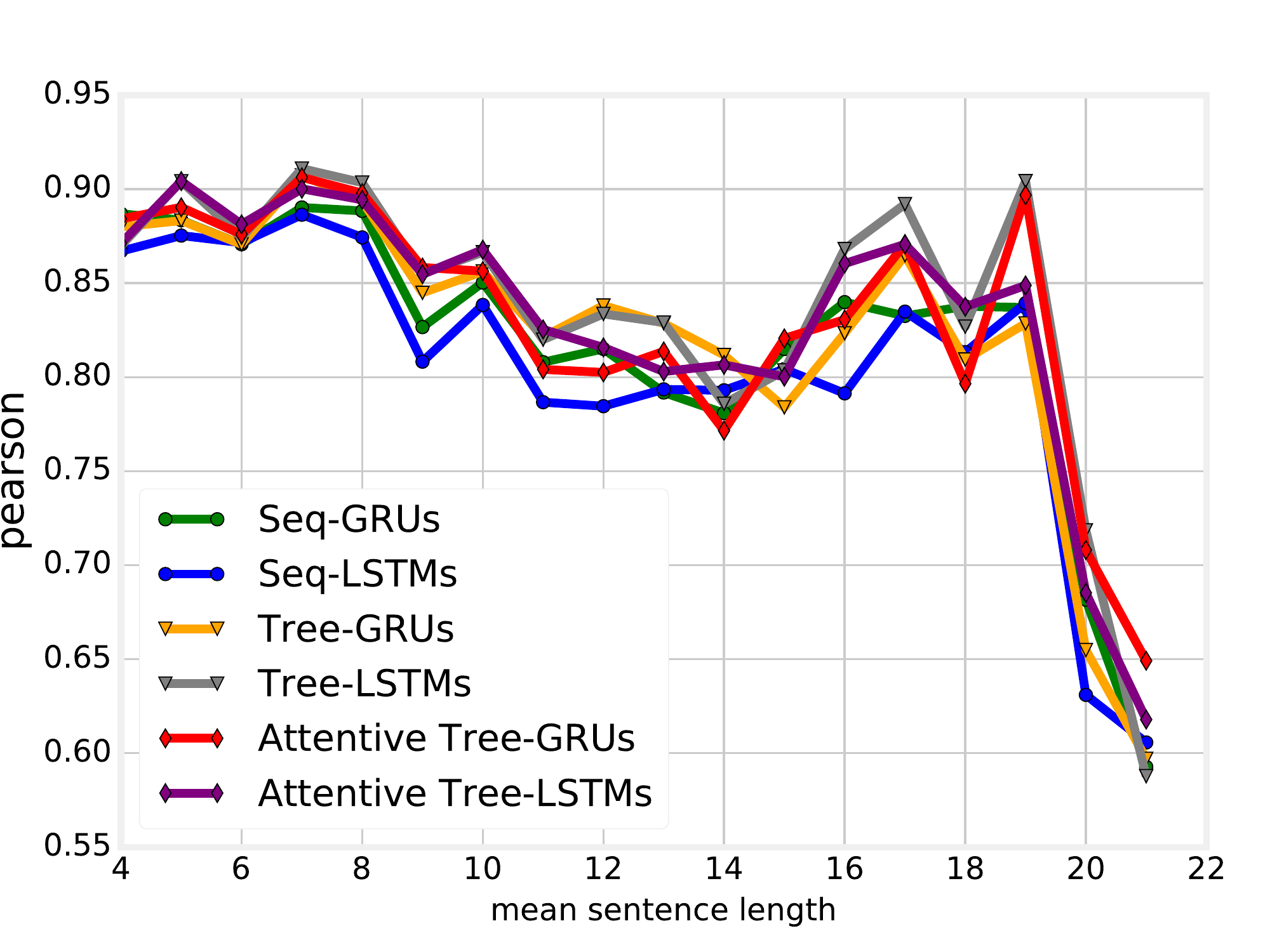}}
\hspace{10px}
\subfigure[MSRP dataset]{\label{stats-msrp} \includegraphics[scale=0.35]{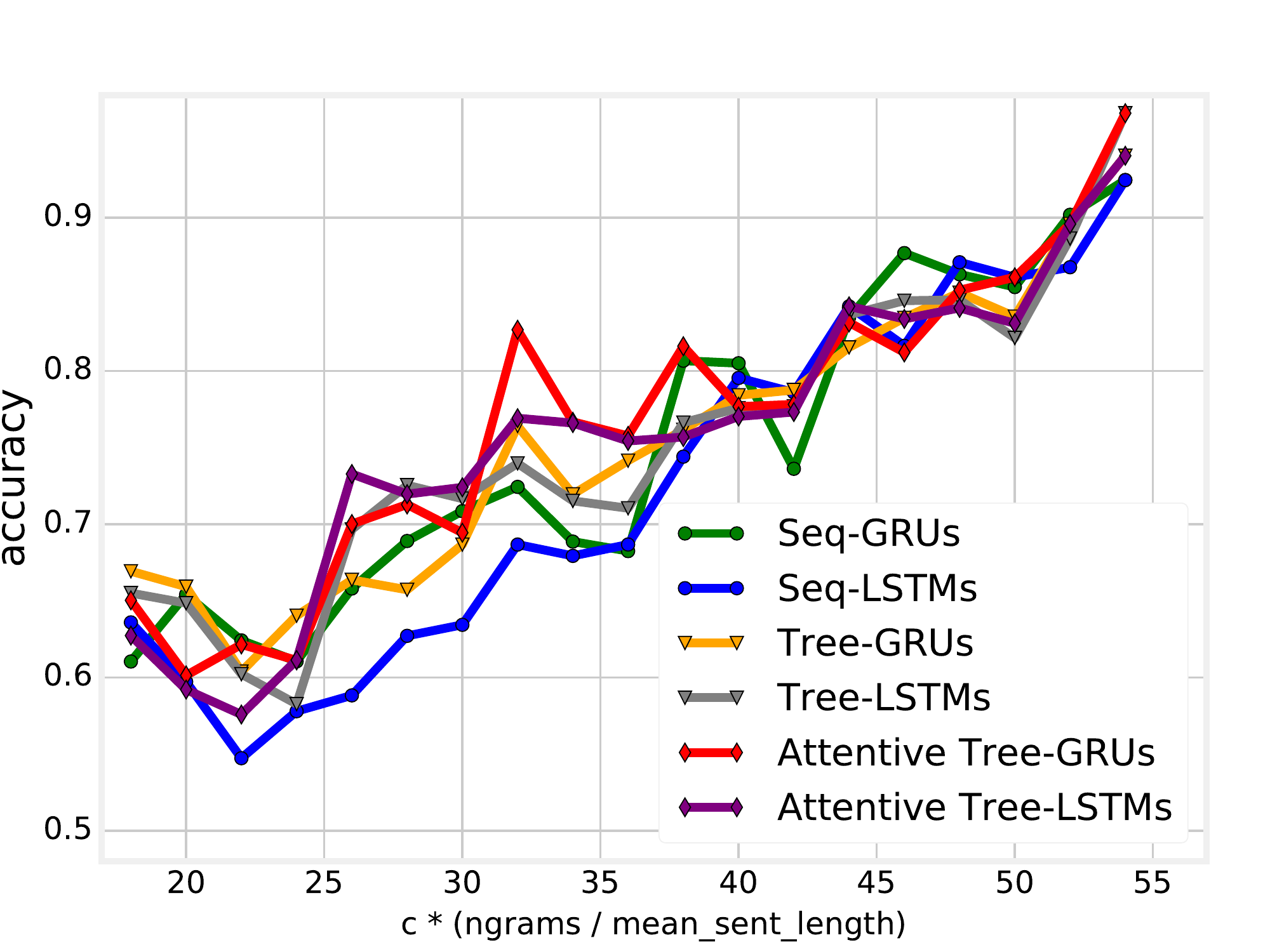}}
\caption{Qualities of different models based on mean sentence length and $n$-grams overlap}
\end{figure*}

\section{Quantitative Analysis}

\begin{table}[htbp]
\begin{center}
\scriptsize
\begin{tabular}{lp{6cm}p{6cm}cc}
\hline

\hline

\hline
\textbf{Dataset} & \textbf{Sentence 1} & \textbf{Sentence 2} & \textbf{GT} & \textbf{Pred} \\ 
\hline
\multirow{3}{*}{SICK} & The black dog is playing with the brown dog on the sand & A black dog is playing with a brown dog on the sand & 4.8 & 4.8 \\ 
& A brown dog and a black dog are playing in the sand & The black dog is playing with the brown dog on the sand & 5.0 & 4.2 \\
& A brown dog and a black dog are playing in the sand & A black dog is attacking a brown dog on the sand & 3.5 & 3.4 \\
\hline
&The study is being published today in the journal Science. & Their findings were published today in Science. & 1 & 1 \\
\multirow{3}{*}{MSRP} &The launch marks the start of a new golden age in Mars exploration. & The launch marks the start of a race to find life on another planet. & 1 & 1\\
&Last year, Comcast signed 1.5 million new digital cable subscribers.&Comcast has about 21.3 million cable subscribers, many in the largest U.S. cities. & 0 & 1\\

\hline
&Sunlight is the nutrient source for some fungi ? & The main difference between plants and fungi is how they obtain energy. & 0 & 0\\
\multirow{5}{*}{AI2}&Sunlight is the nutrient source for some fungi ? & Plants are autotrophs, meaning that they make their own ``food'' using the energy from sunlight. & 0 & 0 \\
&Sunlight is the nutrient source for some fungi ? & Fungi are heterotrophs, which means that they obtain their ``food'' from outside of themselves. & 0 & 0 \\
&Dead organisms is the nutrient source for some fungi ? & Most fungi live in soil or dead matter, and in symbiotic relationships with plants, animals, or other fungi. & 1 & 0\\
&Dead organisms is the nutrient source for some fungi ? & Relate the structure of fungi to how they obtain nutrients. & 1 & 0 \\
&Dead organisms is the nutrient source for some fungi ? & From dead plants to rotting fruit. & 1 & 1\\
\hline

\hline

\hline
\end{tabular}
\end{center}
\caption{\label{results-examples} Example predictions from the test set. \textbf{GT}: ground truth, \textbf{Pred}: predicted score.}
\end{table}

\begin{figure*}[h]
\centering
\subfigure[]{\label{sick-att-1} \includegraphics[scale=0.33]{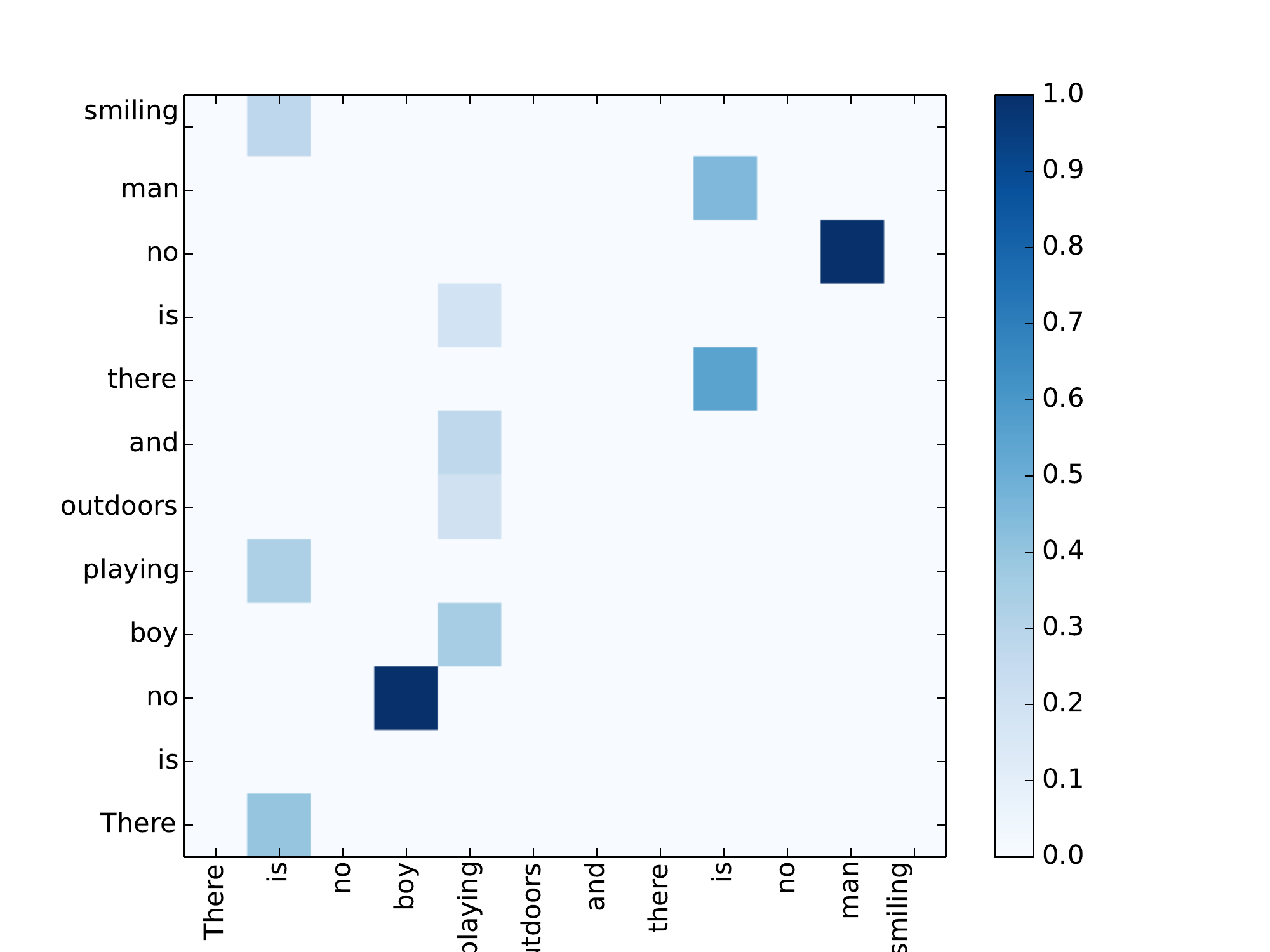}}
\hspace{1px}
\subfigure[]{\label{msrp-att-1} \includegraphics[scale=0.33]{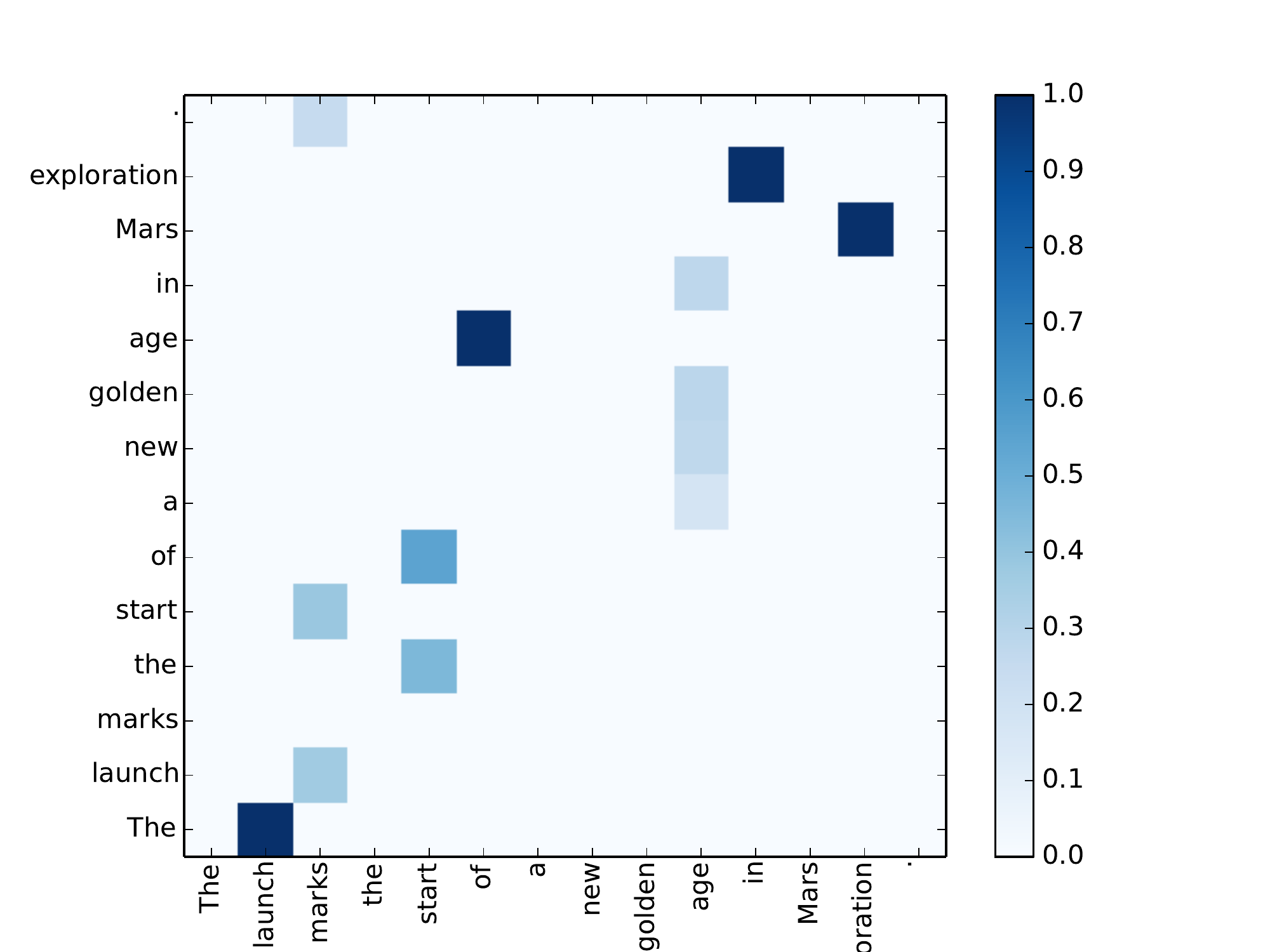}}
\caption{\label{fig-att}Heatmap of attention weights.}
\end{figure*}

\paragraph{Example Analysis}
Table~\ref{results-examples} presents example predictions that are produced by our Attentive Tree-LSTMs. The first group shows that our model is able to predict semantic similarity score nearly perfectly on the SICK dataset. We argue the reason is that the sentences of SICK dataset are image and video descriptions whose sentence structure is relatively simple and there are less uncommon words and named entities in the vocabulary. The second group gives us three examples on the MSRP test set. We find that our model can identify whether two fact statements are paraphrases, but fails to recognize the numbers (in group 2, line 3). We presents the examples on AI2-8grade dataset in the last group. We can observe that our model is efficient to select the false questions, while our model is difficult to select the true answers, unless the evidence of question is very strong.
\paragraph{Effect of Sentence Length}
In order to analyse the effect of mean sentence length on the SICK dataset, we draw the Figure~\ref{len-sick}. We observe that the Pearson score become lower as sentence become longer. Compared with the Seq-RNNs, the Tree-RNNs obtain a little improvements. Specially, the Attentive Tree-GRUs proves to be more effective than Tree-GRUs when the mean sentence length reaches to 20.
\paragraph{Effect of $N$-grams}
In the MSR paraphrase corpus, a hypothesis is that two sentence tend to be paraphrases when the value of their $n$-gram overlap is high. As a result we present the Figure~\ref{stats-msrp}, x-axis is the  normalized $n$-grams overlap whose value is computed by $c * \frac{(unigram + bigram + trigram)}{mean\_sent\_length}$, where $c$ equals to 50, and y-axis is the accuracy. We can observe that the Attentive Tree-GRUs are more effective than Tree-GRUs when the value of normalized $n$-grams overalp is less than 40. The results suggest that our attentive models are more general.
\paragraph{Attention Visualization}
It is instructive to analyse which child the attentive model is attending over when constructing the head representation. We visualize the heatmaps of attention weights shown in Figure~\ref{fig-att}. The words at x-axis are modified by the words at y-axis with a weight (greater than \emph{zero}). For example in Figure~\ref{sick-att-1}, the 5th word at x-axis is ``playing'' whose children are ``boy'', ``outdoors'', ``and'' and ``is''. We can observe that the word ``boy'' holds a higher weight among all the modifiers. It means that the branch rooted with ``boy'' contributes more when constructing the representation of subtree whose root node is ``playing''. This phenomenon is very reasonable because the sentence is describing a image of ``a \textbf{boy} is \textbf{playing} something''.

\section{Conclusion}
In this paper, we introduced a way of incorporating attention into the Child-Sum Tree-LSTM and Tree-GRU that can be applied to the dependency tree. We evaluate the proposed models on three sentence pair modelling tasks and achieve state-of-the-art performance on two of them. Experiment results show that our attentive models are effective for modelling sentence pairs and can outperform all non-attentional counterparts. In the future, we will evaluate our models on the other sentence pair modelling tasks (such as RTE) and extend them to the $seq2seq$ learning framework.
 
\section*{Acknowledgements}
This work was funded in part by the National Key Research and Development Program of China (2016YFB0201900), the National Science Foundation of China (grant 61472459, 61370021, U1401256, 61472453), Natural Science Foundation of Guangdong Province under Grant S2013010011905.

\bibliographystyle{acl}
\bibliography{coling2016.bib}
\end{document}